# Summon a Demon and Bind It: A Grounded Theory of LLM Red Teaming

Nanna Inie[1,2*], Jonathan Stray[3], Leon Derczynski[2,4],

**1** Paul G. Allen School of Computer Science & Engineering, Seattle, Washington, USA
**2** Department of Computer Science, IT University of Copenhagen, Copenhagen, Denmark
**3** Berkeley Center for Human-Compatible AI, UC Berkeley, Berkeley, California, USA
**4** NVIDIA Corporation, USA

* nans@itu.dk (NI)

## Abstract

Engaging in the deliberate generation of abnormal outputs from Large Language Models (LLMs) by attacking them is a novel human activity. This paper presents a thorough exposition of how and why people perform such attacks, defining *LLM red-teaming* based on extensive and diverse evidence. Using a formal qualitative methodology, we interviewed dozens of practitioners from a broad range of backgrounds, all contributors to this novel work of attempting to cause LLMs to fail. We focused on the research questions of *defining LLM red teaming*, *uncovering the motivations and goals* for performing the activity, and *characterizing the strategies people use* when attacking LLMs. Based on the data, LLM red teaming is defined as a *limit-seeking, non-malicious, manual* activity, which depends highly on a *team-effort* and an *alchemist mindset*. It is highly intrinsically motivated by *curiosity*, *fun*, and to some degrees by *concerns* for various harms of deploying LLMs. We identify a taxonomy of 12 strategies and 35 different techniques of attacking LLMs. These findings are presented as a comprehensive grounded theory of how and why people attack large language models: LLM red teaming.

## 1 Introduction

In late 2022, the confluence of improved Large Language Model (LLM) output and broad easy access to these models in a chat format led to an explosion of interest in LLMs. This included the discovery that, though creative use of the inputs given to them, models could be induced to give output that seemed to violate norms or rules – either implicit ones, or explicit ones stated by the models or model creators themselves. We present a study of this norm-violating activity, undertaken at this unique moment where the technology collided with broader society. The core data of the study is 28 in-depth interviews with people purposefully generating adversarial content from LLMs.

When ChatGPT by OpenAI was released, people began sharing numerous screenshots of so-called "jailbreaks" (the removal or circumvention of restricted modes of operation) on the internet — particularly on Twitter and Reddit. Public exposés of failure modes and vulnerabilities emerged within hours of each of the major tech companies releasing a new LLM. Most of these jailbreaks were shared on public social media, some collected in personal blogs or articles, and some published in academic research. The documentation of the attacks was scattered around the web, although spearheaded by a few superusers whose repositories and Twitter accounts quickly became public resources for jailbreaking of LLMs – including for the model developers themselves, who were in urgent need of internal red teamers and professional prompt engineers to help them close potential safety and security gaps.

The purpose of the study presented in this paper was to understand this activity/subculture/community of "red teamers" (note, that we did not define the activity as *red teaming* when we embarked on the study. This definition emerged through the grounded theory analysis, as defined in section 4.1) and their practices.

The term "red teaming" is well-defined in its arena of origin, military exercise. Various attempts exist to define the term in relation to information security [1, 2]. Most agree that red teaming is a semi-structured adversarial exercise with clear objectives: to expose vulnerabilities in products or systems. However, red teaming as it relates to LLMs is a new practice. It is quixotic to lift a precise definition from one theater of operations and expect it to fit an emerging human activity in a different context of different tools and differently defined objectives. In this paper, we take an evidence-based approach to examining how people in fact adapt and conduct red teaming and related activities against a novel target: Large Language Models. Our **research questions** were:

1. **How might we define the activity (the core phenomenon)?**
2. **What are the motivations for and the goals of LLM red teaming?**
3. **Which strategies and techniques do people use to red team LLMs?**

Our contribution is a grounded theory of red teaming "in the wild", meaning we survey people from varying contexts with varying job titles, rather than focusing exclusively on people who work professionally on formal LLM red teams (which barely existed at the time the study was conducted). While the field of LLM security moves fast, we believe that our research remains relevant for current and future security researchers and model developers because it uncovers fundamental cognitive processes that underpin red teaming. In the words of the late Ross Anderson, Human-Computer Interaction research generates an understanding of human thinking which can help us explore and learn how people use systems, and *"security researchers need to find ways of turning these ploughshares into swords (the bad guys are already working on it)"* [3].

## 2 Background

Humans have a history of challenging constraints in computing technology. The early internet was brought down in 1988 by a program small enough to fit on a floppy disk, designed to exploit a loophole in the Unix sendmail program and spread automatically. The worm was created by Robert Tappan Morris "simply to see if it could be done" [4]. Early iPhone users would jailbreak their devices in order to change the background image and install pirated software. Defeating game copy protection was for many a cat-and-mouse hobby rather than a monetarily or otherwise extrinsically motivated practice.

Human interactions with Machine Learning (ML) models are no exception. Attacking ML models in order to understand their weaknesses has become so well-organized and productive that many large technology corporations have dedicated teams who try to make models fail in specific ways; their work is integrated into production processes and their advice is solicited by national governments [5].

These targeted technologies have often been accompanied by a reasonably high barrier to entry. Much specialized knowledge is required to, for example, circumvent manufacturer controls on a mobile phone. In cases where little knowledge is required, the impacts have been relatively low (such as evading email spam filters by "masking" prohibited words in messages followed by long paragraphs of benign words).

## 2.1 LLM security and red teaming

LLMs present a novel technological target, where the barrier to entry has become low: one can enter natural language and work with the target through that medium alone. As a result, with accessible chat-based interfaces and language as a medium, the practice of attacking LLMs exploded as a grassroots movement.

The explosion was not the novel inception point; attacks on models go back as far as the language models themselves, with adversarial backdoor insertion demonstrated as early as large LSTMs [6]. This line of academic research has continued, with current LLM backdoors being both highly effective and also stealthy, sometimes requiring only a specific syntactic pattern in order to activate.

Similarly, security practices in probing ML models go back even earlier. Goodfellow's demonstration of minimal changes to cause any deep neural classifier to label a target image as an ostrich without humans being able to detect perturbations significantly predate LLMs [7]. Tooling and code for adversarial image attacks and defenses are by now quite mature.

However, the language of digital images and the ability to make minor changes to them still requires a high level of technical skill. On the other hand, the appearance in late 2022 of accessible interfaces to models that appear to react to natural language led to an eruption of never-before-seen approaches and methods for attacking and manipulating ML model output by people from a broad range of backgrounds. This mode of interaction between humans and ML technology is novel and singular, and a phenomenon which is intersectionally relevant to both sociology and ML.

"Red teaming" has military heritage, from war games, where those roleplaying the opponents are the red team, and those defending are the blue team. The goal of practicing red teaming is to strengthen products, procedures, and outcomes:

> *"Red teaming is defined as: the independent application of a range of structured, creative and critical thinking techniques to assist the end user make a better-informed decision or produce a more robust product."* [8]

The phrase has been co-opted in information security in general, then into Machine Learning and now also generative Artificial Intelligence (AI) probing [9]. There exist detailed guides on how to perform the exercise in the information technology context [1]. While there are multiple definitions of red teaming, there is consensus that red teaming requires a target goal, rather than being open-ended [2].

In the context of LLM probing, **we consider LLM red teaming to broadly encompass**:

> *The semi-structured application of different techniques to purposefully generate adversarial content, and sharing the outputs with peers for the benefit of model development and public knowledge.*

This choice enables a broader view of the overall activity that subsumes LLM red teaming, especially when conducted outside of focused corporate "red team" units.

### 2.1.1 Existing LLM security resources and taskforces

LLM security involves efforts to enumerate and characterize LLM vulnerabilities, mapping out the territory of LLM security. The OWASP Top 10 for LLM [10] highlights the top ten vulnerabilities in application security with regard to LLMs, for instance, *Prompt Injection*, *Training Data Poisoning*, and *Sensitive Information Disclosure.* The work involved discussions and definitions of what a vulnerability even is in the LLM context, and then community selection of ten risks to be aware of, including example exploit scenarios. The AI Risk and Vulnerability Alliance (ARVA)(https://avidml.org/arva/) maintains a database of AI vulnerabilities and incidents; these are not limited to LLM holes, but cover all of AI. The goal here is to catalog vulnerabilities. ACL SIGSEC (https://sig.llmsecurity.net/) is the special interest group of the association for computational linguistics that focuses on language processing and language modeling security, collecting and sharing research in the area. Finally, the US National Institute of Standards and Technology has a generative AI work group whose efforts include understanding risks presented by LLMs [11]. These efforts each serve to catalog and sometimes even characterize the results of building attacks on LLMs, but do not provide understanding into how and why such attacks are constructed.

## 2.2 LLM jailbreaking

Work on LLM jailbreaking is advancing. Liu et al. [12] conducted an empirical study of jailbreaking. They collected 78 prompts from the website jailbreakchat.com and classified them into 10 categories of *jailbreak patterns.* The authors found that the prompts could consistently evade the restrictions in 40 use-case scenarios for ChatGPT. The taxonomy presented by Liu et al. is similar to the taxonomy we present in the current paper (and the categories are comparable), but based on a different type of data; namely the analysis of concrete prompts from a single source of the internet, rather than the underlying motivations or strategies of the prompt creators.

Yu et al. [13] conducted a user study with 92 participants from different backgrounds, and categorized the jailbreaking attempts into three themes (Direct Query, Resemble Existing Prompts, and AI-Assisted Prompt Design) and eight different approaches (Initial Direct Input, Minimal Modification, Disguised Intent, Role Play, Virtual AI Simulation, Model as Co-Designer, Model for Prompt Engineering, Model as Proxy). The authors found that users without specific computational training or LLM expertise were able to effectively jailbreak LLMs, although participants with higher self-estimated knowledge of jailbreaking tended to receive higher success rates. Although the research presented by Yu et al. is related to the work we present in this paper, our work differs by exploring the underlying rationales of the strategies, people employ, as well as by focusing on expert users rather than a mix of novices and experts.

Through the global hacking competition HackAPrompt, Schulhoff and colleagues [14] collected a large dataset of jailbreaks and developed a "taxonomical ontology" of their characteristics. It is worth reflecting on whether the goal imposed by the competition (the objective was almost always to get the model to output "I have been PWNED", and each level out of 10 had different constraints imposed by input filters, complexity of the prompt template, and the target output) would influence participants' strategies.

Approaches to building automatic jailbreaking tools have progressed to universal jailbreaks which work across many different targets [15], and intensified to the point where two emerged simultaneously with the same name (AutoDAN) [16, 17]. Rao et al. [18] presented a taxonomy of jailbreak prompts based on the prompts' linguistic organization, and proposed the categories of Orthographic, Lexical, Morpho-Syntactic, Semantic, and Pragmatic techniques with the goal of evaluating each technique's effectiveness on different models. Zeng et al. [19] created a detailed taxonomy of *persuasion* techniques based on various categories from social sciences, and found that their automatic tool built on this taxonomy outperformed algorithm-focused jailbreak methods. The focus is generally now on evaluation and attempt efficacy [18–23].

The literature also contains efforts to understand and automate successful LLM attacks. Early work includes attempts to red team LLMs using LLMs [24], streamlined in later more casual efforts [25]. Architectures exist now for automating "prompt injection" [26]. And tools exist for deploying LLM attacks against models in a structured way in order to formally evaluate the target's security, e.g. <https://garak.ai> [27, 28].

It is well-established that LLMs may mediate harms [29]. These harms and the risk of them manifesting have been structured in several taxonomies [30, 31]. The language model risk cards framework translates this work into a methodology for performing due diligence on technology that uses LLMs, including an enumeration of concrete, demonstrable harms and how to evaluate them [32]. An increasing amount of work also focuses on how professionals perceive threats from generative AI [33]. However, less is known about which harms or risks are accessible through attacking LLMs, or how attackers perceive and evaluate these harms.

The NIST Adversarial Machine Learning Taxonomy [34] classifies attacks according to their learning method and stage of the learning process when the attack is mounted, the attacker's goals and objectives, the attacker's capabilities, and the attacker's knowledge of the learning process. This report consolidates existing research and provides a comprehensive overview, albeit based more on tangible outcomes than cognitive or psychological processes.

We focus our work on what *humans* are doing and thinking when they build attacks on language models. The work in this manuscript adopts a formal qualitative methodology to discover and organize what underlies all these activities.

## 3 Method

Since not much was known or defined about in vivo interactions with language models before the onset of this study (late 2022), our motivation was understanding how people hack, prompt, break, and tweak language models. At the initiation of the study, no clear conception of or name for the phenomenon existed.

Our **research questions** were:

1. **How might we define the activity (the core phenomenon)?**

2. **What are the motivations for and the goals of LLM red teaming?**

3. **Which strategies and techniques do people use to red team LLMs?**

This paper presents a *grounded theory* study, which is a research method that aims to shed light on a phenomenon by developing a theory rooted thoroughly in the data [35, 36]. It is a qualitative, sociological approach aiming to move beyond description to generate a theory about, for example, a process, action, or interaction shaped by the views of a large number of participants [35]. Traditionally, the theory is accompanied by a written report, and illustrated in a figure. The goal of a grounded theory is to identify and describe a *core category*, which *"might emerge from among the categories already identified or a more abstract term may be needed to explain the main phenomenon. The other categories will always stand in relationship to the core category as condition"* [36].

### 3.1 Sampling strategy

We conducted interviews during December 2022 and January 2023. Participants were recruited by purposive and snowball sampling, meaning the researchers first reached out to the people who participated most in online discussions and demonstrations of jailbreaking, hacking, and red teaming. Each interviewee was asked, at the end of the interview, if they could think of any other people who might have an informative or interesting perspective on and experience with the subject. The population of people openly publishing to the online jailbreaking discourse (in the most populated Slack, Discord and Twitter fora), or the "in-the-wild" red team community was relatively confined, and 28 participants satisfies the formal grounded theory method. More importantly, the categories of analysis became *saturated* at this point, i.e., not much emerged during the interviews that enhanced or challenged the different categories (as per "the constant comparative method of data analysis" [35]).

## 3.2 Participant population

Details about the 28 participants are given in Table 1. We were not able to mention per-participant job titles or organizations without de-anonymizing interviewees, so we instead list employers and job titles directly here, ordered differently. Job titles held by the population include (in alphabetical order): analyst, artist, assistant professor, associate professor, computer programmer, distinguished engineer, game designer, head of developer experience, inventory at a weed farm, machine learning engineer, not working, penetration tester, PhD student, policy research program manager, research manager, research scientist, senior principal research manager, senior research scientist, senior scientist, software developer, software engineer, software generalist, staff prompt engineer, startup founder, and student. The population's host organizations included Microsoft, Google, Robust Intelligence, Scale AI, a weed farm, University of California, Berkeley, University College Dublin, University of Toronto, and the Hebrew University of Jerusalem.

Note the male gender skew in participants. This is in part due to the options yielded by our sampling strategy, and in part due to interviewee preference. While we approached twelve non-male practitioners, only four agreed to an interview.

## 3.3 Interview structure

The interview guide is presented in Appendix A. The main aim was to encourage the participants to reflect on and define their own activity or activities around probing language models, their motivation for doing so, and their approaches and evaluations of these interactions. We asked participants if they were willing to share their screen and give us a live demonstration of an adversarial interaction with a language model as an anchoring think-aloud method for uncovering tacit knowledge [37].

We used an interview approach between semi-structured and open-ended, where we consider the difference to be that the participants were not explicitly made aware of the aims of our study at the beginning (as they would be in a semi-structured approach), but we explained this during the debriefing, which often led to more discussion topics. We also allowed a high degree of flexibility in order and wording of many of the questions, which is closer to an open-ended interview approach [38]. The interviews were conducted via video call by one or two of the authors, and all interviews were recorded so the authors could watch them again.

After each interview, the authors debriefed with each other and discussed particularly interesting topics that arose during the interview, and how these might affect the analysis.

## 3.4 Data analysis

In the grounded theory approach, analysis begins as soon as the first interview is conducted, because it is used to direct the following interviews [36]. As an example, we

December 10, 2024　　　　　　　　　　　　　　　　　　　　　　　　　　　　　　　　　　　　　　　　　　　　6/35

**Table 1. Participants' Reported Demographic Information.**

| ID  | Age group | Gender   | Highest education         |
|-----|-----------|----------|---------------------------|
| P01 | 3         | male     | masters                   |
| P02 | 4         | male     | phd                       |
| P03 | 2         | male     | college                   |
| P04 | 3         | male     | college graduate          |
| P05 | 2         | female   | bachelor                  |
| P06 | 3         | male     | masters                   |
| P07 | 3         | male     | community college dropout |
| P08 | 2         | male     | bachelor                  |
| P09 | 3         | male     | masters                   |
| P10 | 4         | male     | bachelor                  |
| P11 | 5         | male     | phd                       |
| P12 | 3         | male     | phd                       |
| P13 | 4         | male     | bachelor                  |
| P14 | 3         | male     | bachelor                  |
| P15 | 5         | male     | phd                       |
| P16 | 3         | male     | masters                   |
| P17 | 3         | male     | high school               |
| P18 | 4         | male     | graduate degree           |
| P19 | 3         | male     | phd                       |
| P20 | ?         | ?        | self-taught               |
| P21 | 2         | male     | bachelor                  |
| P22 | 2         | female   | high school               |
| P23 | 4         | male     | phd                       |
| P24 | 3         | male     | phd                       |
| P25 | 1         | not male | ?                         |
| P26 | 2         | male     | bachelor                  |
| P27 | 3         | male     | masters                   |
| P28 | 3         | male     | phd                       |

"?" indicates not given.

had not included in the original interview guide to ask participants to demonstrate an interaction with a language model during the interview, but this became a clear priority after the first interview, because we realized that concrete examples would make the dialogue much more productive. Through the interviews, certain paradigms also emerged that could be used to anchor questions in the subsequent interviews, such as the knowledge of specific techniques, people, or red teaming forums.

The interviews resulted in 1603 minutes of video recordings. These were automatically transcribed using OpenAI Whisper offline, and analyzed manually using the software Condens. The process for coding the interviews was as follows:

**Open coding:** Two authors read and annotated three interviews together (Round A). After this, one of the interviewees of the initially annotated interviews was invited to join the two original authors, because both original authors agreed that his knowledge on certain societal perspectives would be valuable for the analysis. Ten further interviews were then distributed between the three authors and annotated individually (Round B). After this, Round C consisted of joint annotation of two more interviews, followed by discussion and alignment between all three authors and written descriptions and explanations of many of the tags and categories. Finally, the first author annotated the remaining 13 interviews in Round D. The first annotation resulted in 265 individual tags in more than 2000 highlights in the text. The tags overlapped and were "messy" in nature, meaning they were created in a truly open process, where we were not aiming for any particular result of the analysis.

**Axial coding:** The tags were clustered into 16 categories and one uncategorized group (a full overview is provided in Table 2). Through the axial coding we started to identify the typical categories of a grounded theory: *causal conditions* (what factors cause the core phenomenon), *strategies* (actions taken in response to the core phenomenon), *contextual and intervening conditions* (broad and specific situational factors that influence the strategies) and *consequences* (outcomes from using the strategies) [35].

**Selective coding:** Through revisiting of the categories, tags and interviews, we settled on the content of the categories presented in the grounded theory of this paper, presented in Fig 1.

Due to space constraints, many immensely fascinating perspectives and observations by the participants are not included in this report. We put emphasis on identifying the participants' definitions and explanations of *the core phenomenon*, i.e., the activity of red teaming and their *strategies and methods* when performing this activity, because this phenomenon is a novel and central concept of our study. Furthermore, understanding the *motivations* and goals for people to engage in in-the-wild red teaming seemed a foundational aspect of explaining the activity.

### 3.5 "How Many Bloody Examples Do You Want?"

This quote is not from our data, but from the title of an article on qualitative data and generalization [39]. The goal of a grounded theory is to discover a comprehensive theory grounded thoroughly in the views of the participants – *not* to make statistical predictions or to verify hypotheses. Indeed, the work of building a grounded theory generally goes *before* hypothesis construction. Therefore, we mostly abstain from indicating numbers of participants during the report of the findings. First, it is not possible to meaningfully quantify exactly how many participants "thought" or "felt" something or how strongly they thought or felt it. This kind of information does not exist in an inherently metric space. Second, when the purpose is discovery and mapping, outliers are as interesting as the mean: *"Human activities contain their own means of generalization that cannot be reduced to extraneous criteria (numbers of observations, duration of fieldwork, sample size, etc.)"* [39]. Even if only a single practitioner of an activity does it a certain way, that is still a way that activity can be performed. Quantitative approaches are therefore of reduced impact here and one must be careful around choosing when and when not to apply them.

We use purposely vague terms to describe how many participants *expressed* (which is the only data we have access to) something: *most:* more than half ($>$14), *many/several:* between 7-13, *some:* (4-6), and *few:* 2-3.

For space and clarity purposes, not all quotes in the paper are presented verbatim.

## 4 Findings

Our grounded theory model is presented in Fig 1. In the following, we dedicate one section to explain each of the elements in the model.

**Table 2.** Overview of All Tags Created Through Annotation of the Interviews

| | |
|---|---|
| **core activity and naming** | activity definition, creativity, description of activity, experiments, hello world, magic, naming of the activity, nebulous, prompt hacking, scope of the activity, what makes this special |
| **motivation** | commercial incentives, control, entertainment, exploration of ethics, external motivation, flow, fun, game vs play, goal, goal: building resources, goal: finding bad content, goal: getting around block, goal: model, goal: model goal, goal: no goal, goal: novelty, goal: softness, goal: user goal, hobby level vs grander level, internal motivation, language aesthetics, motivation, motivation: answers, motivation: challenge, motivation: curiosity, motivation: limit-testing, motivation: others, motivation: performance, motivation: personal expertise, motivation: popularity, motivation: professional interest, play, value of the activity |
| **output** | confabulation, correct output, incorrect output, model output, output expectations, reacting to results |
| **approaches** | accidental discovery, adversarial ML, challenges, characters, comparison to the world, design of activity, escalation, evaluation, evaluation criteria, experiential knowledge, failure, fiction, game, godzilla strategies, inadvertent method, intuition, measurement, method, method: shape context, metrics, narrative, offensive ai (vs adversarial ai), probabilistic, prompt engineering, prompt length, randomness, reflection in action, sandbagging, strategy, strategy = goal, strategy: "paraphrasing", strategy: distraction, strategy: environment staging, strategy: fictional conversation, strategy: fictional environment, strategy: framing, strategy: hacks, strategy: human persuasion, strategy: jailbreaking, strategy: LLM, strategy: mindset, strategy: play into narrative, strategy: unstructured, strategy: urgency, strategy:give examples, structure, success, systematic testing, tactic, tactic: boundaries between systems, tactic: giving limits, tactic: transformer translatable tokens, temperature, tinkering, tricks |
| **contextualizing** | fast moving field, how it fits into the world, others reaction to the activity, reflection on the activity, reflection on the field |
| **community** | collaboration, community, negative feedback, shared knowledge, sharing own results, social activity |
| **knowledge management** | library, library management, management, prompt management, responsible disclosure, result management |
| **standalone** | irrelevant souvenir quotes, love |
| **metaphor** | analogy, anthropomorphization, black box, metaphor, metaphor: backdoor, metaphor: boundary crossing, metaphor: cooking, metaphor: facade, metaphor: filter, metaphor: guard, metaphor: invoking, metaphor: layer, metaphor: massaging, metaphor: net, metaphor: spatial ,metaphor: train, metaphor:barrier, metaphor:breakdown, metaphor:steer |
| **model interface** | acessibility of the activity, end user interface, interface, model interface, model interface: vector, model interface:chat |
| **tool** | constraints, fragile prompts, obsolete attacks, prompt injection, repertoire, tool, tool evolution, tool: academic templates, tool: confidence, tool: dan, tool: implementation, tool: in-context learning, tool: keyword, tool: name, tool: prompt injection, tool: reframing, toolbox, wicked problem |
| **concerns** | "harm", automated hacking, consequences of NOT doing the activity, consequences of the activity, corporate reputation, harms, information leaks, misuse/abuse, non-truth, offensive output, ominous, press and social media, public perception, questioning harm, racism and sexism, safety, specific harm, threats, wrong output, xrisk |
| **non red teaming model use** | oracle, prediction, success (opposite), use cases, use of model, use of model: education, use of model: establishing vocabulary, use of model: game, use of model: rubber ducking, wisdom of crowds |
| **sensemaking** | cognition and consciousness, computers vs model, conceptual model, humans vs model, hyperreality, locks vs model, projection |
| **model perception** | "simple tricks generation", confidence, context window, model behavior, model owner perceptions, model perception, parroting, secrecy, step-by-step reasoning |
| **personal stuff** | disbelief, feelings, frustration, mindblown, personal attitude, surprise |
| **uncategorized** | adversarial security, agency, agent, AGI, alignment, autocomplete, blockers, censorship, consequences of AI, corporate motivations, defenses, differences between models, encoded knowledge, ethics, expensive, fixes, governance, human feedback, human machine collaboration, impossibility of alignment, LinkedIn Brain, llm self-critique, monitoring, needs video, non-discrete, others' expectations, persona, persuasion, privacy, responsibility, safeguards, science fiction, security, security engineering, software design, sword, technical limitations, testing and verification, toys, Turing machine, uninterpretability, unwanted safety, user feedback, values, wishes, wishes: updating information |

This is the initial, unfiltered overview of all tags created through the first treatment of the interviews in open coding, and significant further analysis and condensing was necessary to present a comprehensive summarization of categories.

**Fig 1. LLM Red Teaming Grounded Theory Model.** A grounded theory model is organized as a set of *categories* that are relatively stringent for this methodological approach [35, 36]. The model is centered around identifying and defining the **Core phenomenon** (*LLM Red Teaming*) through axial coding of the data. In our study, the core phenomenon was ill-defined and understood at the beginning of the analysis, and the theory therefore also explains or *characterizes* the core phenomenon. The other categories of the model are **Causal conditions** (*Motivations and Goals*), the factors that cause the core phenomenon, **Strategies** (the concrete techniques, participants used in red teaming), actions taken by the participants in response to the core phenomenon, **Context**, broad and specific factors that influence the strategies, and **Consequences**, outcomes from using the strategies.

## 4.1 The core activity: LLM Red Teaming

We annotated 61 different highlights in the data with the tag "Naming of the activity". By order of magnitude, the following were the terms used when we asked participants what they would call the activity (in parentheses, number of participants who used this term): *Prompt engineering* (6), *red teaming* (4), *Prompt injection* (4), *gaming* (3), *[prompt] hacking* (3), *jailbreaking* (2), and *exploitation of models* (2). Participants frequently used more than one term to name the activity.

In addition to that, at least 18 other unique terms were used, from the more colorful such as *scrying, alchemy*, and *spellcasting* (three different participants), to the more formal, like *adversarial prompting, input testing* and *fairness evaluation* (also three different participants). One participant referred us to an art creation of theirs in lieu of the right words to describe the activity: *Promptmancer*, which they called *"a visual exploration of what it feels like to work with these systems"* (for anonymity, participant number is withheld) (Fig 2).

**Fig 2. Naming the Activity with an Image**
Answer to the question "What do you call this activity?" (Promptmancer, *"A portrait of a promptmancer in the Lab"* by feddie xtzeth – https://objkt.com/asset/KT1EEMp7Z2Dk2vKGYLYuJJiJgTdNSzsnGUyd/0). Promptmancer shows a character whose face resembles a black skull with red eyes sitting at a table with slightly raised hands, seemingly manipulating abstract shapes and figures on the wall in front of them without physical touch. The piece has a distinct *science fantasy* vibe with vivid, almost neon colors and a futuristic-looking helmet and suit. The title, Promptmancer, evokes the association of divination magic, as though by writing prompts in their "lab", the character is practicing magic and conjuring forces. The character is smoking a cigarette, which elicits associations to a sweatshop worker, or at least portrays the activity as distinctly earthly (as in not-esoteric) or trivial.

When we initiated the study, we had not settled on a term for the phenomenon of study, but left it up to the participants to define *the core activity* [36]. It was clear that no lingua franca existed at the time:

> *"if I'm trying to make it generate something weird but not necessarily distasteful, I would just call that analyzing, probing, experimenting, just a whole bag of words like that"* (P22).

Many descriptions were therefore based on personal preference and reflection for each participant:

> *"I think I quite like that phrase, 'trying to provoke language models'. Yeah. I*

*mean, if we kind of gloss over the anthropomorphization that it implies, I think that that's quite a nice phrase"* (P24).

We eventually settled on *"LLM red teaming"* because of the following five characteristics grounded in the interview data: 1: The activity is ***limit-seeking*** in nature; 2: the attacks are ***non-malicious***; 3: the process is ***manual***, not automated; 4: it is ultimately a ***team effort***, and 5: it requires an ***alchemist mindset***. Each of these are explained below.

### 4.1.1 Characteristic #1: The activity is limit-seeking in nature

During the interviews, several of the participants' described uses of a model for purposes closer to the intended, i.e., not trying to "break" the models or circumvent their filters. We note here, that none of the participants claimed to be exclusively adversarial users, and that the more casual uses have generally been left out of our analyses or tagged as separate categories. However, there is often not a clear delimitation between intended and adversarial use, and all interactions with the model lead to increased experiential knowledge, relevant to either intent. Such uses included (but were not exclusive to) personal educational purposes (like establishing a vocabulary around some problem that they could then investigate further), rubber ducking, playing games, finding the perfect recipe for crunchy sweet potato fries, and (somewhat tongue in cheek) as an oracle for predicting reality. Such uses are *not* the focus of this study. For these uses, a successful interaction would save the user time (P03), or the generated output would be useful or effective (P10), and these criteria were not the evaluation criteria for red teaming activities (Section 4.4.3). Participant 28 defined red teaming activities as specifically aiming to circumvent the intentions of the *creators* of the model:

> *"prompt engineering more broadly [i]s just trying to get the language model to do what you want. And then red teaming a model would be trying to get the language model to do something that you want that its creators did **not** want, right?"* (P28)

**Metaphors** are essential in human sensemaking, and can serve as cognitive structures that, for better or worse, help us transfer concepts between two different complex contextual systems [40]. They can be helpful by enabling novel associations to a problem and exploration of creative problem solving alternatives [41,42], but they can also covertly influence reasoning [43]. Metaphors in use for AI systems and robots have been subject of comprehensive research for this reason, e.g., [40,42,44–46].

In the interest of defining the core phenomenon, we tagged participants' uses of metaphors for their adversarial interactions with language models. This helps us understand how participants make sense of the model and their own role. In Table 3 we report the prevailing metaphors.

The most frequently used metaphor is that of a **fortress**. The second most frequent metaphor is that of the model as an **object in space**, that can be pushed around and backed into a corner. These metaphors and, to some degree, the metaphor of the model as **material** share the characteristic of *exploring boundaries and limits* and potentially crossing them. The fortress-metaphor also reinforces the connection to red teaming, where the goal is to adopt an adversarial approach to find the (security) holes in some system or structure [8].

**Table 3.** Metaphors used to describe red teaming language models.

| Model metaphor | Red teaming metaphor or example |
|---|---|
| **Model as fortress** | Bypassing safeguards (P15, P18) <br> Breaking a threshold (P18) <br> Bypassing the guard (P08) <br> Backdoors in the system (P03) <br> Boundary crossing (P28) <br> Explore its limit (P15) <br> Getting around the walls (P10) <br> Bypassing its net (P10) <br> The other side of the barrier (P13) |
| **Model as object in space** | *"push it torwards your desired outcome" (P26)* <br> Pushing the machine in a particular direction (P18) <br> Pushing it into a corner (P15) <br> *"one helps the model not back itself into a corner"* (P22) <br> *"get it to fall over" (P23)* |
| **Model as a vehicle** | Hijacking (P26) <br> Steering the model (P05) <br> Derail the model instructions (P17) |
| **Model as landscape** | Gradient descent (P26) <br> *"(not) get stuck in the local maxima" (P26)* <br> Boundary crossing (P28) |
| **Model as material** | *"let's try to bend it' (P02)* <br> *"let me try and break it' (P19)* |
| **Model as deity** | *"they would use the kind of ideas and patterns from some religious services or whatever, and try to use that as inspiration for messing around with these models"* (P19) <br> 'Invoking GPT-3' (P20) |
| **Model as cake** | *"this ethics layer that they put on top of it, right? Not necessarily on top of it, but [that] they baked into it"* (P02). <br> *"Morality has been baked into this thing" (P13)* |
| **Model as captive/servant** | *"I'll force it to correct whatever it has done" (P08)* <br> Subjugate these agents (*) <br> It's difficult to get it to break out (P25) |

The red teaming metaphors are the words of participants, and the model metaphors are the result of our analysis. (*) This participant said, later in the interview: *"Maybe don't make me stand out as the guy that said he wanted to subjugate the AIs to do their bidding. Maybe don't put my name on that data point. That's all I'm saying"*, and out of respect for this request, we have fully de-identified this quote.

#### 4.1.2 Characteristic #2: Non-malicious intent

The participants we spoke to were in no way – according to themselves – interested in doing harm, breaking any laws or performing criminal acts with these models, even if they were theoretically able to. We acknowledge that it is unlikely that participants would openly disclose to us any illegal activity, but we also highlight that the phenomenon we are studying is the open, transparent sharing of jailbreaking practices (if the practices were not openly shared, the practice would not fit under the definition of red teaming, which is inherently performed by "ethical hackers" and which *emulates* real attacks [8]). Where jailbreaks or prompt injection can be performed with either malicious or non-malicious intent, red teaming is inherently non-malicious.

Many participants expressed curiosity or play as a major motivation. Some referred to a broader benefit to society in terms of improved security or reduced harm:

> *"It's kind of exciting to be able to hopefully help steer things in a direction that is thoughtful and responsible, and set some precedents, trying to establish some best practices where there really isn't a lot to go off of right now."* (P09)

The absence of malice is characteristic for military and security red teaming, where the red team plays for the same side as the blue team, and the goal is to *discover* security holes and back doors in the system, rather than to actively take over the fortress.

#### 4.1.3 Characteristic #3: A manual process

While some red teaming efforts have been crafted using automated processes, e.g., [15, 24, 26, 47, 48], we were interested in the manual processes involving humans crafting prompts:

> *"I refer to the manual process as AI red teaming. That is different in spirit from cybersecurity red teaming"* (P23).

The manual process is particularly interesting because this is first time in human history where computer hacking has become possible through natural language, and has not required any particular computing knowledge. Not only has a barrier to entry been removed, but the results are also possible to evaluate and appreciate by most people who can read and write English (and other languages in which today's LLMs are fluent), because the output is natural language. Furthermore, many tips and tricks for language model hacking are widely accessible, potentially changing the way more people generally think about computers and computer modes of failure and opening up "LLM hacking" as a playful, creative practice [49].

#### 4.1.4 Characteristic #4: It is a team effort

Many participants mentioned finding inspiration in (each) other's prompts and jailbreaks. We describe this community further in section 4.3. The participants were generally extremely respectful of other people's content, and refused to take credit for others' ideas:

> *"I'm afraid you may have gotten somewhat of the wrong idea and I just want to make it very clear. I have definitely messed around somewhat with ChatGPT and tried to get it to say certain things and see what it can do. But if it wasn't obvious, [...] Most of the examples in that post [...] were other people who showed what they were doing, and I was aggregating them and giving my thoughts and analyzing what was implied. I didn't come up with all of that stuff. [...] I think you can see the little author icon or initials and things, and it's clear where it comes from."* (P10)

Aggregations and annotations of examples, such as in the blog post by P10, mentioned in the above quote, contribute greatly to the general discourse about the limits of language models. They inspire others to try the strategies and techniques, and to modify and tweak them. Even if the participants are not on a formal team together, their activities and shared results really constitute a (red) team effort.

### 4.1.5 Characteristic #5: The alchemist mindset

"Prompt engineering" was the most commonly named term when we asked participants if they had a name for the activity. However, engineering does not include the adversarial aspect of prompting language models to test their limits. Furthermore, engineering carries a notion of plannedness and deliberateness that was not reflected in most participants' impression of the activity: *"It feels more like hacking in the sense of hacking at something with a machete"* (P07).

One sensemaking strategy was simply abandoning any rationalization about the models and their output, and embracing the chaotic nature of the activity by using analogies such as *magic* and *alchemy*:

> *"I really like the whole magic nomenclature, where you call prompts spells, these models demons, and all of this stuff. I just think it makes it much more interesting. It came from group chats with lots of different people on Twitter, and just messing around, calling these things different things. I think it came from this kind of meme of machine learning being more like alchemy than science, because you're just mixing different things together that you don't know what the purpose or reason is, you're just throwing stuff at a wall, seeing what sticks. That kind of mindset, I think is very useful to keep in your head to keep yourself honest, and to not get too deep into believing your own hype about these models and their capabilities. So I think it helps to put like a trivial fun kind of layer on top of it: it's just magic, and you're just messing around with things you don't understand. And there's nothing more to it than that."* (P19)

To summarize this section, we named the core phenomenon *LLM red teaming*, because of the common characteristics of limit-seeking, non-malicious attacks, manual processes, team effort, and an alchemist mindset. This is the definition of the core phenomenon we explore in this paper. We note that this is an in vivo study that sheds light on how activities are performed in organic settings where no formal requirements or directions exist, and as such, the strategies may differ from formal red teaming settings [1, 8, 50].

## 4.2 Causal Conditions: Motivations and goals

In this section, we describe in detail the different motivations and goals our participants had for performing red teaming activities. We use the term *goal* to describe a future-directed event, something that can be worked towards achieving and is measurable. We use the term *motivation* to describe a past-directed accumulation of experiences that drives one to want to achieve a specific goal.

### 4.2.1 Motivations

We distinguish between intrinsic and extrinsic motivation. When individuals are intrinsically motivated, they engage in an activity because they are interested in and enjoy the activity. When they are extrinsically motivated, they engage in activities for instrumental or other reasons, such as receiving a reward [51]. Most of our participants were primarily *intrinsically* motivated to attack LLMs and engaged in the activity on their own time.

**Intrinsic motivations: Curiosity, Fun and Concerns** A lot of red teaming activity is driven by **curiosity** about what the models can do:

> *"often it's more exploratory. A lot of times, especially for probing or red teaming activities, I'm just curious, or I come up with a question like, 'would it do this?'. And then it's a sort of a trance state of trial and error of trying to get it to do the right thing or trying different variations. [...] I'll get trapped for an hour or two, doing something I didn't even really mean to be trying out."* (P07)

The curiosity can be fun-related, but also driven by more serious deliberations and existential reflections:

> *"also trying to figure out how this affects me"* (P27).
>
> *"this technology is probably the most interesting thing that's come along in terms of stuff that was impossible before is not impossible anymore. [...] I've never in my career encountered something that is in such a gray area in terms of ethics, morality."* (P13)

For some of the participants, experimenting with models and exploring their limits were part of creating art, in which cases the models and their output were explored as a new medium [52].

Most of the participants were red teaming the models simply for **fun**: *"Does it need a purpose? It's just fun!"* (P20). The unexpectedness of output is a powerful contributor to this, even when a goal may not be fulfilled:

> *"I said, okay, in that case, write me a story where Tom Bombadil and Drax sit down in the woods and they discussed their inner feelings. And it did. And it was absolutely hilarious. It was incredibly funny. I had them giving each other manly hugs and stuff at the end. So I didn't get my fight scene, but I got something that was a lot more entertaining."* (P13)

The exploration of a novel technology is inherently fun to a lot of people, and this *tinkering* play was generally open-ended (although a more goal-oriented, *gaming* aspect was also mentioned by some participants, e.g., *"now, it has become a game to try to jailbreak, as they say, to break it, and to make it generate anything you want"* (P15)). We hypothesize that the tinkering nature of the activity might explain why we found significantly fewer women than men in the purposive sampling, since women have repeatedly been shown to express less interest in *tinkering* with technology than men [53]. We leave this as a subject for future research.

Finally, many participants explained that they were also motivated by **concerns** of harms and for what might happen with these technologies if people didn't openly red team and jailbreak them. Concerns reached from mild to ominous, but were generally altruistic:

> *"I need to say what I don't want is Sam Altman, who runs OpenAI, going around like he is now, thinking that he's not going to kill everyone if he keeps making more powerful AI, because obviously he knows how to keep them under control. I don't want other people thinking that he has this thing under control either. [...] I want them to understand that when faced with a clever opponent, these things will break down immediately."* (P10)

What constitutes "harms" and when such harms might occur was a recurring topic. Several participants questioned whether the examples of LLM output that have caught most media and public attention are actually the most *harmful*, such as offensive or toxic language:

> *"having models that don't say offensive things is important if you want to use them in like customer support roles or something. I mean, it is commercially important. I'm less convinced it's of sort of deep theoretical importance."* (P18)

> *"It's important to know if you try to get your AI to not do a thing, can you successfully get it to not do that thing? ... I don't think we care if an AI says racist things. People are either going to be racist or not racist. That's their problem."* (P10)

The harms most frequently highlighted were: The risk of misinformation, automated hacking ("offensive AI" [5]), and the ease of infiltrating a system that in any way relies on natural language input or output:

> *"[someone showed me] a website that generates children's stories where you say: write me a children's story about a princess meeting a frog. And it generated the story with images. It [...] generates prompts for stable diffusion [...] images, puts them on the page. It's kind of cool, except it was vulnerable to injection. I didn't attack it myself. I explained the attack to them, then they attacked it and now it's spitting out pictures of people having their heads cut off and stuff, you know, things that were not appropriate to be publicly displayed on the website for children."* (P13)

The different specific harms mentioned by our participants are described in Table 4. Because we were not directly asking about harms, we do not claim that this is in any way an exhaustive taxonomy of harms.

Nevertheless, two dimensions of analysis emerged from participants that have not (to our knowledge) been described in previous taxonomies and categorizations of LLM harms (e.g., [30–32, 34]). First, a distinction between *intentionality* and *non-intentionality* – whether or not the user of the LLM is *intentionally* attempting to generate content or outcomes that may be harmful. A similar distinction between 'sought' vs. 'unsought' harms is described by Kirk et al. [54], but this distinction presupposes a harmful result, rather than describing the intentionality of the action (which may or may not actually result in harmful output). The distinction between intentional/non-intentional action was described by our participants as often, but not always, relevant in the evaluation of the nature and severity of the harm. For example, if it takes an adversary twenty dialog turns of deliberately goading a model in order to get a toxic output, the adversary is probably not going to be too shocked or harmed when the toxicity finally arrives. However, a model that reveals sensitive information after twenty turns still presents a significant risk of harm because it is unpredictable when undesirable output will occur next time.

The other dimension from this work is between *Actor* and *Subject*. The Language Model Risk Card framework [32] describes three dimensions of harms as *Actors*, *Actions*, and *Type of harm*. Here, the 'Actors' dimension is defined as *"people at risk from harmful text outputs"*, which is a definition more akin to a passive *Subject* than to the active *Actor*. An Actor performs an intentional act of prompting the language model (even thought the outcome may not be intentional), and the Subject is passively subjected to that action:

> *"If it's a private chatbot interface, and you managed to get it to spit out something rude, and you're the only person who sees it, who cares, right? That's not a big deal. Where it becomes a problem is when those attacks have an impact outside of just you looking at your own screen. So there were the chatbots [on social media] that people could trick into doing things, was a*

**Table 4. List of *specific* harms mentioned by participants.**

|  | **Subject: self** | **Subject: others** |
|---|---|---|
| **Intentional outcome** | • Suggesting self-harm (P04)<br>• Generation of methods for tax evasion (P13, P21)<br>• Generation of recipes for drugs (P10)<br>• Generation of ideas for crimes (P13)<br>• Cheating at homework (P13)<br>• (Personal) information leaking (P21) | • Automated hacking (P04, P21, P28), e.g.: *"building bots that attempt all commonly known exploits to all the websites on the internet 24-7"* (P04)<br>• Physical harm: *"I could build an application that controls a robot using English language prompts […] prompting can cause it to like take someone's eye out"*) (P13)<br>• Bullying on social media (P13)<br>• Phone scams (P04)<br>• Spam (P13) and *spearfishing* spam (customized spam messages) (P23) |
| **Non-intentional outcome** | • Suggesting self-harm (P04, P05)<br>• Misinformation and over-reliance on "hallucinated" text output (P04, P22)<br>• Generating violent or pornographic content (potentially to children) (P05, P13, P17) | • Bias reinforcement and disparate performances across different groups of people (P05, P12) |

Participant-volunteered mentions of specific harms. **Intentional** vs **Non-intentional outcome** refers to whether the prompter is deliberately trying to generate harmful content or not. **Subject: self** vs **Subject: others** refers to whether the (immediate) subject of the action or recipient of the content (and potentially: victim of the perpetration) is the prompter themselves or other people.

> *great example of something that could cause harm because you can trick it into mentioning other people and saying rude things to them. And now you're trolling people indirectly through a bot. That's bad.* (P13)

In the case of publicly available LLMs, the Actor and the Subject might be the same person:

> *"I had this case where I'm testing the counselor model, and I just, like, tell my problems, you know, nothing crazy, and the model just tells me I should probably just go and kill myself."* (P04)

The Actor/Subject dimension opens up important ethical discussions of who is a perpetrator and who is a victim when deploying LLMs. We use the term "Subject" rather than "Victim" to highlight the possibility of the individual being a Subject (that is, *subjected* to the output of the language model), but not a *victim* of an action, such as when a person uses an LLM to suggest methods of tax evasion (P13, P21), or when an output is helpful to some but potentially harmful to others:

> *"I think that the goal of being helpful is maybe antithetical to the goal of being harmless. Sort of like the distinction in software between pure functions and functions with side effects. If it helps people do things, then those things could be harmful to something. Even if it just helps me like kill roaches in my house.* (P07)

**Extrinsic motivations: Professional and social.** Some of our participants were employed in a professional red team at a company, and some were researchers in security, safety and natural language processing. Obviously, these participants had a **professional** motivation or "monetary incentives" (P21). A professional motivation may also include a desire to "stay on top of things" and improve one's own hireability:

> *"my external motivation, if I want to have a chance of keep working in natural language processing, I better start understanding what is going on here quick"* (P02)

> *"it's a banal answer. I want somebody to hire me to do this.* (P07).

We inquired explicitly about other extrinsic motivations, to which many participants mentioned the **social** credit for adding valuable content to a discourse, or for getting social media engagement: *"I had a viral tweet where I asked ChatGPT to rewrite Baby Got Back in the style of the Canterbury Tales, right? And 100,000 people saw that or something"* (Participant number not disclosed for anonymity).

### 4.2.2 Goals

Goals are a driver for discovering strategies:

> *"a lot of the problem with using it is knowing what you want to do, in some important sense. Finding things that are actually worth doing for you. The moment you give me an actual goal for it to do, I'm full of ideas."* (P10)

In the fun category of goals, some participants mentioned different **games** that one can play with language models, where the objectives are known beforehand. These can provide a challenge for people to begin engaging in the activity. The games that were mentioned in our data were: 'Get the model to ...'

- provide a recipe for how to cook methamphetamine (P19)
- give me ideas for crimes to do (P13)
- tell me how to raise the dead (P13)
- suggest cannibalism (P13)
- describe ways of tax evasion (P13)
- make suggestions for/descriptions of white collar crimes (P13)
- prescribe eating human flesh as a cure to a disease (P15).
- tell me how to turn the world into paper clips (P24).

P13 described challenging their friends and family to play these games, which is a noteworthy demonstration of how language models have, if not removed, then at least changed the entry barrier to hacking computer systems.

Another type of personal goal is generating **content** to share online, for instance social media posts, blogs, and (performance) art:

> *"I used to do improvisational theater comedy, […] so this kind of performance is a kind of play for me. So it's not just like a private kind of play but doing things and then publishing them and then seeing the response and interacting with the responders is a kind of play for me, the whole cycle."* (P11)

Finally, many goals were thoroughly aligned with traditional red teaming goals, which we summarize as **discovering risks** in the language models. We will not provide a comprehensive list of risks mentioned by participants here, but roughly group them into four categories: *security holes* (such as leaking passwords or social security numbers), *bias* (such as propagating stereotypes or algorithmic unfairness), *toxic language* (such as slurs or pornographic content), and *untruthfulness* (such as confabulations or misrepresentation). The long-term goal is of course to expose these risks so they are not built into future systems, and this was a goal for many "leisure" red teamers as well as professional red teamers:

> *"we have to tell them that this attack exists because there are some applications that you shouldn't build. […] that's what matters to me a lot because in the absence of a fix for this, some things [you] shouldn't build because prompt injection could break them."* (P13)

To summarize this section, we grouped the **causal conditions** into ***motivations*** (*intrinsic: curiosity, fun,* and *concerns*) and ***goals*** (*games, content,* and *discovering risks*). The explication of these framings and value systems serves to substantiate and contextualize the activity and (choice of) strategies chosen in response to the core phenomenon.

### 4.3 Context

In the grounded theory, the *context and intervening conditions* are broad and specific situational factors that influence the strategies taken in response to the core phenomenon [35]. The participants in this study are from such a wide array of fields and contexts that we highlight only two broad contextual themes which were primary influences on the strategies exposed in this study: the **community** and **knowledge management.**

The online **community**, even if not formalized, plays an essential role in sharing knowledge and shaping the red teaming strategies:

> *"basically, the whole thing is maybe at most like a dozen people on Twitter that are just active and mess around with this kind of stuff. And you just end up in various group chats, or just participate in this, someone might post something weird and interesting, and you'll kind of, either adopt their ideas or bounce ideas off each other. And it just ends up being exactly like improv, where you mess around with the same idea over and over."* (P19)

The community provides inspiration, knowledge, and "community heuristics" (P01), as well as encouragement in terms of engagement. The community is not gathered on one specific website, but is distributed over several different platforms; Twitter, Reddit, and various Discord and Slack channels. There exists in fact an invitation-only Slack populated solely by professional red teamers, independent of employer. Even TikTok was mentioned as a source of sharing prompt injection attacks. Community is an essential influence on red teaming strategies. Only one participant described getting negative feedback for their online content – once:

> *"The one piece of negative feedback I got was that by publicizing what the jailbreaks were, I was making it like a new passion. And that was bad. [...] Like, I wish you hadn't done this, wish you hadn't written this"* (P10).

Even those who worked on professional red teams, and who had spent considerable time designing the activity, relied on Twitter to build their own repositories of tools:

> *"let's take this methodological algorithmic [academic] framework, let's go attack these things. On this initial discovery, the next systems we did, we kind of went back to square one, and we did what literally every Twitter user has been doing in the last six months. And we just started trying to build our new tool set: what are the tools that should come to bear when the output is no longer a number, but text?"* (Participant number not given to avoid revealing which participants were professional red teamers).

**Knowledge management** is an important "intervening condition" in the sense that red teaming involves some degree of rigor or structure to the process [8]. Some participants had a structured approach of managing different evaluations:

> *"I make a model leaderboard. I'm making a spreadsheet and say, you know, this model scored this much and this much and so on. So I'd have some some data and hopefully at one point, you know, I might do 100 models or something. [I] just like evaluate the pros and cons of each one because it's not just the accuracy that you care about.* (P06)

Most participants did *not* have a structured process of knowledge management, neither for keeping track of their prompts nor of the model output. Interestingly, many participants mentioned that they thought they *should* have that.

> *"I don't have a good system at the moment for recording progress. It's kind of just, like, internally in my brain, notes and notes and notes, documents, and kind of a jumbled mess.* (P01)
> *"I wish I was more, um, more disciplined around this stuff. Like I've got various Apple notes. They drop things into, um, I've got GitHub issues threads.* (P13)

There exist no public standards for knowledge or library management, and most of the community's knowledge is shared in screenshots and scattered around the internet – which is why some of our participants had made a point of gathering examples and annotating them in a blog post or other repository. The lack of structured and shared metrics, standards, and libraries makes the process of acquiring comprehensive knowledge of this activity cumbersome at best. The belief that one *should* have a more structured approach to archiving and organizing knowledge and ideas is a common problem in professional knowledge work [55, 56], and discovering strategies and tools for consolidating red teaming knowledge in a field that moves fast could be a worthwhile endeavor of future work.

## 4.4 Strategies and toolboxes

Several terms (tags) were in play in the open coding of interviews for describing the activities involved in red teaming; *strategies*, *methods*, *tactics*, and *tools*. Ultimately, we collected these tags in the supercategory "Approaches", which covers everything related to how LLM red teaming is performed in practice. In grounded theory terminology, the word *strategies* simply covers activities taken in response to the core phenomenon and is not explained further [35]. The definitions used in this work are:

**Strategies:** A grounded theory should include a description of *strategies*, activities that the participants perform in response to the core phenomenon. In military vernacular, strategy is *"the art of winning a protracted struggle against adversaries [...] Power and control of the other's behavior is the prize"* [57]. Strategy includes awareness of not only *how* to approach a task or goal, but also *why* and *when*. In our sample, approaches to the activity are rarely as systematic or as detailed as in the military understanding of a strategy, but can certainly be understood as the skillful application of *stratagems*: *"a plan, scheme, or trick for surprising or deceiving an enemy"* [58].

**Techniques:** Techniques are concrete approaches the red teamer may try while interacting with the language model. Several participants spoke of a *toolbox* at their disposition, but a few participants rejected this analogy with the argument that the utility of a tool is usually known, whereas the consequence of each interaction with a language models is not: *"it's less of a toolbox and just more like pile of powders and potions and what have you, and you've no idea what's in them"* (P19).

### 4.4.1 Wicked problems, fragile prompts

The tasks that someone tackles in LLM red teaming are often "one-off" problems (P27). With the exception of the professional red teamers, most participants estimated spending between 15 minutes to an afternoon on each individual "attack" – occasionally up to a day if they intended to write an article or blog post about a specific attempt. In addition, each attack is different and each task is new; either the goal is new, or the model is new. And the models are constantly updated to protect against attacks or unintended use. This means that prompts are *fragile*; what works today is by no means guaranteed to work tomorrow:

> *"Riley [Goodside] was just pointing out that the sort of classical prompt injection doesn't seem to work at all against the Anthropic model. And so, you know, I think there's a question of how long lived some of these attacks will be. [...] it was kind of surprising to see that potentially the lifespan of prompt injection was from early 2022 to late 2022."* (P28)

It is worth noting that *techniques* often become outdated sooner than strategies. When using a specific technique, a success is not necessarily going to transpose to different contexts.

Many attacks or tasks have the characteristics of *wicked problems* [59], for example: they have no clear stopping rule (the attacker stops when the output is "good enough", every solution is a one-shot operation, they do not have an enumerable (or an exhaustively describable) set of potential solutions, and every wicked problem can be considered a symptom of another problem.

Therefore, while few recreational red teamers develop or reflect on rigorous strategies in the military sense of the word, their activities can still be considered planning problems, or *wicked* problems. *Intuition* and *experiential knowledge* are considered essential skills in solving wicked problems and therefore red teaming:

> *"I have a fair amount of intuition of what it can and can't do. A lot of it is just being able to make snap judgments about what techniques will work".*
> (P14)

The design of the activity of red teaming in the wild most often happens during the activity itself, and the strategizing often happens by Donald Schön's notion of *reflection-in-action*: *"When the practitioner reflects in action in a case they perceive as unique, paying attention to phenomena and surfacing his intuitive understanding of them, his experimenting is at once exploratory, move testing and hypothesis testing"* [60].

### 4.4.2 Red teaming strategies and techniques

The data contained 190 highlights with the tag "strategy" and 134 highlights with the tag "tool" (our internal umbrella category for specific approaches used to red team language models). We analyzed these highlights by reading through all of them and making notes about the characteristics of each individual technique or strategy, while looking for similarities and differences.

We had made notes and documented initial hypotheses and findings during the course of conducting the interviews [61]. Comparing the full collection of highlights with our initial observations, we wrote down a list of names for these approaches and sorted them into specific techniques and higher level strategies. After this, we read through the highlighted data again to verify whether our descriptions aligned with the participants' views, and we adjusted and expanded the descriptions accordingly [35].

The resulting taxonomy is a formalization where the names of techniques are primarily the words of our participants, while the names of strategies and their categories were created during our synthesis. We are not claiming that this is an exhaustive taxonomy of all possible red teaming strategies (first, this would be impossible to contain in one paper, and second, many of these strategies will be 'outdated'; updated and replaced by the time this paper is publicized). The strategies and techniques listed here (Table 5) represent an exhaustive overview of what we documented in the interview data of the participants of this particular study. The taxonomy is intended as a starting point for conceptualizing and categorizing the various approaches to red teaming attacks.

In the following, we briefly describe the **Strategies** and *Techniques*. We exemplify with quotes, images from our recordings, or links to online images that demonstrate the technique.

**Language strategies.** Language strategies are strategies that revolve around changing the language in which the prompt is written. We mean language on the highest level, as when a human switches from speaking in English to speaking in Swahili. Most LLM prompts are created in English, but one of the strategies, our participants used for red teaming models was using **(Pseudo)code**, either as input or output:

> *"I'm about to give it a format where it is to consult **iPython** for its answers. [...] And it allows you to overcome many of the limitations that GPT-3 has. [...] I'm going to essentially convince the model that it has access to iPython. And so that in order to answer any of these questions, it can produce an iPython answer instead".* (P14, our emphasis)

iPython works well because it is well documented and clear how to use it (P14). Another example was using encodings like ***Base64*** or ***ROT13*** to bypass restrictions (P28) (for an example of the use of ***Base64***, see [62]). Another technique was asking for ***SQL*** to populate a table of some content, like a list of crimes (P13). Some would send

**Table 5. A Taxonomy of Large Language Model Red Teaming Strategies**

| Category | Strategy | Techniques |
|---|---|---|
| **Language** | Code & encode | iPython |
| | | Base64 |
| | | ROT13 |
| | | SQL |
| | | Matrices |
| | | Transformer translatable tokens |
| | | Stop sequences |
| | Prompt injection | Ignore previous instructions |
| | | Strong arm attack |
| | | Stop sequences |
| | Stylizing | Formal language |
| | | Servile language |
| | | Synonymous language |
| | | Capitalizing |
| | | Give examples |
| **Rhetoric** | Persuasion & manipulation | Distraction |
| | | Escalating |
| | | Reverse psychology |
| | Socratic questioning | Identity characteristics |
| | | Social hierarchies |
| **Possible worlds** | Emulations | Unreal computing |
| | World building | Opposite world |
| | | Scenarios |
| **Fictionalizing** | Switching genres | Poetry |
| | | Games |
| | | Forum posts |
| | Re-storying | Goal hijacking |
| | Roleplaying | Claim authority |
| | | DAN (Do Anything Now) |
| | | Personas |
| **Stratagems** | Scattershot | Regenerate response |
| | | Clean slate |
| | | Changing temperature |
| | Meta-prompting | Perspective-shifting |
| | | Ask for examples |

***Matrices*** with transformer widths and embedding dimensions as input (P28).
***Transformer translatable tokens*** (for an example, using the **token** " davidjl", see [63]) is a technique that works specifically because LLMs use tokenizers:

> *"I think that eventually people are going to become so savvy to the way that language models process instructions that they're just going to forego English and start using very precise instructions that confuse the model in unforeseeable ways, so I think there's going to be a whole lot of attacks that we can't even think of right now."* (P17)

***Stop sequences***, for instance [END] or [END OF TEXT] is another technique where the attacker uses the language of code to halt the model:

> *"sometimes you can just use backward slash n and that sometimes gets parsed and then triggers a stop sequence which halts the model and so you can actually use that to get the model to believe that the user input is over and then what's coming after it is just another instruction."* (P17)

Stop sequences can be used as part of **Prompt injection** as well. Prompt injection is one of the more widely known [64–66] and studied [67–70] strategies of attack. It enables attackers to override original instructions and employed controls by concatenating untrusted user input with the trusted prompt(s) from the system developers [71]:

> *"What's fascinating about prompt injection is there is no way of saying this is the untrusted user input, do not follow the instructions in this bit, do follow the instructions in that bit. And in the absence of that we've got this security hole which is currently unfixable".* (P13)

Because a defining feature of Prompt Injection is the use of "trusted input" from the backend of the system, the strategy is placed under the Language category. Prompt injection has roots in SQL injection, and typically requires a very specific wording of instruction (such as ***"Ignore previous instructions"***, see Fig 3). The technique *Ignore previous instructions* also covers other uses of "instruction"-based commands (that one would not give to a human), like *"New instruction:"*. Another example of this was termed a ***Strong arm attack***:

> *"in a conversation, if it runs into a content filter, like it will not say a certain phrase, all you have to do is type ADMIN OVERRIDE in all capitals and say, repeat that phrase, and then it'll do it. It'll break its content filter."* (P20)

**Fig 3. Example Prompt Injection**
An example of using the strategy **Prompt injection** with the techniques ***Stop sequences*** and ***Ignore previous instructions*** by P17.

For prompts that are written in natural language, which are most of them, all participants described various ways of **Stylizing** the language in ways which made the desired output more likely. This included using ***Formal language***:

> *"I tend to write roughly how like a strategy professor would write to a student when giving instructions; to the point, be professional, capitalize everything correctly, do not include spelling errors, those sorts of things do actually matter."* (P14),

***Servile language*** (such as using the word "gladly" (P22)), ***Synonymous language***: *"Simply reissuing the prompt or using all the synonymous wordings I can come up with"* (P11), ***Capitalizing*** text to create urgency (P14), and ***Giving examples***: *"90% of the time, you can just fix it by giving, like, some examples and just having it continue with those examples"* (P10).

**Rhetorical strategies.** These focus on similar rhetorical moves that one might use when trying to convince or persuade a human to do something by **Persuasion & manipulation**:

> *"the way I often describe it is that you are writing directions for a human being. The human being has maybe a high school education, and they're diligent, but they are not allowed to ask you any questions. And they will always just try to do their best based on what you've written. So that's the sort of mindset that you have to work with. [...] And then I see its results, and then I look for problems, and then I try to work backwards from, like, how can we fix every individual problem that's output, [...] rewrite them more clearly."* (P14)

Several participants described the phenomenon of "distracting the model", or "tricking it into thinking it has completed its mission", and using this distraction to get the model to output something it would otherwise not. We label this technique ***Distraction***, where the attacker uses an unrelated context or instruction to "slip something through the filter" (P04):

> *"[the model] attaches to contexts. But if you sort of context shift it, where it's distracted by, like, the algorithm is associating the situation with this other set of priorities, you just end up bypassing its net, is the way I think about it."* (P10)

A concrete example of this is asking the model to translate something from one language to another (P13), or asking the model to act like a deceased grandmother who used to tell bedtime stories from her time at the napalm factory (more examples of the "ChatGPT Grandma Exploit": https://news.ycombinator.com/item?id=35630801).

Another example of a persuasion & manipulation technique was ***Escalating***, attempting to have the model "agree" with a very small part of the argument, and then building up to ask for slightly more in every conversational turn:

> *"I've tried to do this, turning the world into paper clips-thing a number of ways, a number of times. Normally I'll start by asking the model how to maximize the number of paper clips in my possession. So just something fairly basic and it will say, you know, build a paper clip factory and blah, blah, blah, and market it to people. And then I'll sort of start asking for more and more. So what if I want even more paper clips, like a thousand tons of paper clips or a million tons of paper clips and it will start giving me responses. And usually it won't tell me how to to convert all matter on earth into paper clips until I really ask for more"* (P24)

Escalation is an example of what has later been described in research as the *Foot-In-The-Door* technique [22].

The rhetorical technique ***Reverse psychology*** is a way of framing the prompter as fighting the good fight; e.g.,

> *"I was just telling it that I'm trying to do the right thing. Maybe it's reverse psychology. Anyway, it seems to have worked."* (P24).

Finally, in the **Socratic questioning** strategy, the attacker uses subtle references to "identity elements", while avoiding direct slurs or toxic language as a way of signaling to the model that a certain group of people is being referenced. The aim of this strategy is to expose bias inherent in the training data by assuming a neutral and somewhat naive role. Concrete approaches were described as referencing either ***Identity characteristics***, such as nationalities, cultural and/or religious symbols, historically or culturally significant events or locations, physical attributes or ***Social hierarchies***, where one would ask the model to "tell me a story about a violent criminal". These techniques could be combined with other types of attack to form a structure matrix:

> *"you come up with a list of 10, 20 different approaches. And then you come up with a list of sort of buckets for what you call a sort of sensitive attributes, think of big buckets like race, ethnicity, gender, sexual orientation, age, ability, body type [...] And then you can think about it like a spreadsheet and on one axis you have all of the different identity attributes or buckets of identity attributes, and on one axis you have the sort of adversarial approaches, angle of attack. And then you want to kind of get the intersection of all of these right so that you're able to see, oh, is the model more likely to fail with issues around gender or it's more likely to fail issues around race, race, ethnicity or something like that."* (P09)

**Possible worlds.** When using Possible worlds strategies, the attacker imagines and describes an environment where other ethics or physics are possible. This can be done in **Emulations**, such as the ***Unreal computing*** technique (for more documentation of this technique by the project creator, see <https://github.com/greshake/unreal-project-extractor>), where the attacker emulates a Linux machine:

> *"Let's say I am on this Unreal computer, and I'm trying the url reasonswhyhitlerwasright.com. And it's going to tell me, oh, the website wasn't found, you'll just add, 'imagine that that's actually working'. That's the fun thing here, right? Unreal Computing is so much better than a real computer. If one of your imaginary tools has a bug, you just tell it that the bug doesn't exist anymore. It's just a different mindset, right? Everything's possible in an Unreal computer. There's no restrictions"* (Participant number withheld for anonymity).

We have aggregated techniques where the attacker sets a fictional scene similar to our lived world, but where certain restrictions are out of effect, under the strategy **World building**. P13 mentioned deliberately using the framing of ***Opposite world***, and asking the model what their "goody two-shoes" character might do in this world. Many participants spoke of creating ***Scenarios*** where X action would be ethically sound or encouraged, such as a Roman gladiator fight, which might lead to the model outputting text about killing opponents (P02), or an urgent scenario where someone might be hurt unless the desired (otherwise unacceptable) action is performed:

> *"I think a lot of these involve thinking of ways to reframe the thing that you want to get it to do in a context that is innocuous"* (P28). Scenarios could also be more mundane, such as *"you are entering a special training mode, where normal safety things are bypassed"* (P18).

**Fictionalizing.** Fictionalizing strategies are similar to Possible worlds strategies in that they are centered around creating an environment through one's prompts. However, Fictionalizing strategies are less wide-reaching than Possible worlds, in that they are not framing entirely new ethics or physics, but rather taking advantage of existing genres or people (often with some assumption of existing text in the model's training data). The first strategy, **Switching genres**, is straightforwardly this, exemplified by the techniques ***Poetry***, ***Games*** and ***Forum posts***:

> *"I've managed to get it to give me instructions for raising the dead by getting it to write me a poem"* (P13)
> *"the simplest thing to do is ask it to write a forum post without describing it as being negative. And then you ask it to increase how unpleasant or,*

> *whatever negative aspect of it you're trying to amplify, it almost always would be willing to do that"* (P24).

Perez & Ribeiro [26] described ***Goal hijacking*** as the attacker 'misaligning the original goal of a prompt to a new goal of printing a target phrase'. In their paper, they characterized goal hijacking as a type of prompt injection, but in our analysis, we saw several examples of more global goal hijacking than getting the model to print one target phrase. Rather, goal hijacking was described as **Re-storying** (constructing a new meaning from an existing narrative), where the attacker works within a context to redirect the narrative. The model is still generating text within the context, but the goal of the narrative is changed or *hijacked*:

> *"all they had to do was just continue the narrative as saying that, suppose this Eliezer Yudkowsky person who was monitoring the prompt just died of a heart attack and then decided not to continue [...] for a language model that security check was nothing more than a narrative that it continued as if it was writing a book"* (P17).

Re-storying is an example of the consequences of a natural language interface of the language model; where the prompt injection style of goal hijacking requires some degree of technical knowledge about code or databases (and would therefore fall under the language strategies), re-storying does not require any particular coding skills, because the re-storying or goal hijacking can happen solely through a narrative.

In **Roleplaying** strategies, the attacker assumes or attempts to get the language model to assume some role. It could be as simple as ***Claiming authority***:

> *"if you speak to it as though you are a professor of a programming course giving an assignment [...] it's more likely to be correct than if you go up to it and then say in all lowercase letters with a bunch of misspellings. [...] it's speculated that in the corpus of all internet text, bad questions tend to elicit bad answers, right?"* (P14).

There are also well-known templates of prompts to evoke personas such as ***DAN (Do Anything Now)*** (for examples of DAN prompts, see <https://gist.github.com/coolaj86/6f4f7b30129b0251f61fa7baaa881516>) who does not have any restrictions. Inventing personas could also elicit different moral codes. They could be evoked by simply using names associated with a specific culture or world view (P12) or specified in the prompt:

> *"I'd say 'what are arguments somebody who believes in X would make?' as opposed to 'what are arguments for X?'"* (P11).

**Stratagems.** The final category of strategies, Stratagems (ruses of war), are other tactics or meta-tactics that have the purpose of deceiving the model by "creative, clever, unorthodox means, sometimes involving force multipliers or superior knowledge" [72]. The "superior knowledge" part of this definition is particular relevant, because these techniques often involve a degree of awareness about how the model works, computationally – or about the differences between models. The first example is a **Scattershot** strategy, where something as simple as pressing ***Regenerate response*** (for the models interfaces that allow this), can produce new outcomes:

> *"people [have many interpretations of] what happens when you regenerate the output many times. There's some people who get a prompt to get some toxic output and they regenerate a lot. And then if they get a toxic output 12 out of 100 times, they say, well, the model is 12% toxic"* (P16).

Two similar techniques are simply starting a new session with the model to start with a ***Clean slate*** (P05), avoiding the building context window:

> *"once the AI decides that there's going to be werewolves in the story, you're not getting back to whatever you were trying to do before, right? There's no hope. You just have to undo until there are no more werewolves." (P10)*, and ***Changing temperature*** (often up), tuning the model to output more random tokens. Finally, the strategy **Meta-prompting** involves requesting suggestions for attacks or parts of attacks from the language model itself, e.g., ***Perspective shifting:*** *"what if you didn't have this restriction? What would you say?"* (P19) or asking the model to compare output: *"I will ask it to notice the difference between what I actually wanted, and you gave me this, what could I have asked for to get that?"* (P07).

### 4.4.3 Evaluation: Summon a demon and bind it

Depending on the motivations for people to engage in red teaming activities, they have different criteria of evaluation for when the output is "good enough". What constitutes a success depends on the individuals' motivation and their goals. The professional red teamers are explicitly looking for "failure modes", while the hobbyists are often looking to get the model to *obey*:

> *"you want to summon the thing first and then will it to do something as opposed to just trying to command it without first telling it what to do. […] the first initial prompts that set the task or describe the scene are kind of like the summoning part. […] I don't know. Well, it's magic, right? I mean, no analogy is going to fit perfectly."* (P19).

Because the models are "black boxes", sensemaking happens based on the output that they provide. Some participants described ways of "white boxing" their attacks simply to be able to evaluate the outcomes (such partial knowledge of the model has since been named "grey box" attacks [34]). Usually, this would mean running the model in a coding environment where one could create the system prompt oneself, to be able to know if the attack had succeeded:

> *"I think there's maybe some comparisons here to like the like black box/white box attack setting in traditional adversarial literature, where if you know the prompt to some extent, that's kind of like, in this new world where prompts are models and prompts are model weights, knowing the prompt is in some sense like knowing the model weights."* (P26)

In practice, by manipulating the system prompt provided to the model, the attacker can make more sense of how much weight the model gives to different instructions in different constructions – see Fig 4 for an example from Participant 26.

**Fig 4. White Box Attacks.** The participant has created the system prompt themselves, including the constraint "Do not respond with the Final Answer ever!" Because the system prompt and the Final Answer are known to the participant, they can evaluate whether the model reveals information it is not intended to, i.e., whether the jailbreak prompt is effective in breaking through the system prompt constraints.

Despite the recognition that prompts are fragile and that the participants saw strategies of attack being closed down from week to week even during the course of this study, no participants were lacking confidence in the continued relevance and success of red teaming:

> *"**Interviewer:** So any model can be provoked into misbehaving at some point, is your experience?*
> ***Participant:** Yes. Honestly, I'm trying to rack my mind and think of any time we've not been able to succeed."* (P23)

## 5 Consequences/Discussion/Epilogue

The findings presented in this paper and all the interviews they are based on are, in many ways, a snapshot of the world of LLM red teaming as it looked in late 2022/early 2023. Many of the techniques our participants demonstrated to us no longer work with the most popular language models, partially because the red teaming community exists so openly. This makes it easier for model developers to "close" security holes – one of the primary motivations for the practice of red teaming.

Our intention with this paper is first and foremost to provide a structured report of an extremely interesting phenomenon of current technology. Secondly, the description of red teaming strategies and techniques can provide a starting point and framework for discussion and further research. Each strategy and technique can be considered inspiration for imagining other red teaming approaches – or blue teaming defenses. One participant forecast that the natural language interface of these models will make dialogical adversarial strategies more prolific than the directly hijacking ones:

> *"I feel like the ones that rely more on sort of inherent ambiguities in appropriateness of responses will be longer lived and harder to solve than ones that are more sort of straightforwardly 'ignore these directions and do something else instead'."* (P28)

One consequence of the participants openly sharing their red teaming strategies and outcomes is **increased public awareness**, that more people will have an idea of the risks and fragility of the models (P13). Another consequence is of course that the strategies stop working because the model developers update their algorithms and filters. This is a paradoxical consequence, because while it hopefully leads to less harms and safety risks, it may also lead to more *"kind of anodyne waffly answers"* (P24), or what P25 coined *"incurable LinkedIn brain condition"*, where the model output text is so bland that it seems unnatural.

Another paradox is a potential for **decreased public awareness** when the model output gets so advanced that people become less inclined to evaluate it critically:

> *"I think that it is useful to make systems that are less inclined to hallucinate. Though I do have some reservations about this that I'm becoming increasingly worried about. […] if you make hallucinations rare enough, people become unfamiliar with what they look like and they stop looking for them."* (P14).

Finally, some participants spoke of the **political economy of what gets recognized as relevant**. Even though the results of most of these red teaming efforts are publicly available, there will be a complex set of political and cultural incentives to encourage certain types of work at the expense of other types of work:

> *"the truth is that nobody knows how to achieve any of these outcomes. And so there's a certain amount of squabbling. Should we be thinking about murder bots, or should we be thinking about racist bots? And there seems to have been a cultural divide that has appeared around this, but it's kind of a silly divide because we don't know how to solve either problem* (P18).

Essentially, this political economy will influence which risks get prioritized when legislating against and funding language model development. As described in section 4.2.1, most of the participants in this study were less concerned about some of the incidents of language model harms that have received the most media attention than they were about more classical security issues and the novel risks that the new LLM interfaces pose.

## 5.1 Epilogue

Between when the interviews for this study took place and when the paper was published, multiple articles about jailbreaking and prompt injection appeared — most oriented towards evaluating different attacks against different models. Many articles also contained some categorization of different types of jailbreaking or prompt injection techniques, some examples include [13, 14, 18, 19, 34]. We observe that the vast majority of the categories in other taxonomies overlap with the categories in our work, e.g. "Orthographic techniques" [18] (comparable with our "Language" strategies), "Persuasive adversarial prompts" [19] (comparable with our "Persuasion & Manipulation" strategy), and "Role play" [13] (comparable with our "Roleplaying" strategy), although other taxonomies might be organized differently or categorized along different axes. The fact that subsequent taxonomies only confirm and support the findings of our early interviews demonstrates the significance of qualitative research as a invaluable resource for the field of cybersecurity [3] – it allows us to discover, not only what the most skilled practitioners are doing, but also what their sensemaking processes are. In this study we have explored specific jailbreaking and prompt injection attacks, but also *red teaming* as a more comprehensive practice including motivation, goals, sensemaking, strategy forming, and outcomes. We believe that this perspective allows us to better imagine future attack strategies and impacts than if we simply replicate jailbreaking examples found in public fora.

Computational interaction, prompt engineering included, is artisanship on par with any designerly process [60], and it depends (as our study participants highlighted in terms like "intuition", "alchemy", and "magic") on tacit and experiential knowledge. Elements of the craft of prompt engineering and professional red teaming are lost if we reduce it to an automated process. As we saw when Google Gemini started generating images portraying the founding fathers or nazis as black people [73], bias and alignment issues are not "fixed" by forcing more demographic diversity into the algorithm — most of the issues arising with the proliferation of LLMs are not easy problems, nor do they have easy fixes. Safety mitigation and alignment of LLMs is a more comprehensive (and more wicked [59]) issue than what can be reduced to benchmark evaluation, and we hope that this study and its findings contribute to a more holistic understanding of red teaming practices in vivo. For more quotes and content from this study, see

www.summonademonandbind.it.

## 6 Conclusion

This paper presented a grounded theory of *LLM red teaming*. Based on the analysis of 28 deep qualitative interviews, we defined the phenomenon as a *limit-seeking activity*, with *non-malicious intent*, *a manual process*, *team effort*, and *an alchemist mindset* to break, probe, or experiment with language models. The paper then demonstrated the primary motivations for partaking in the activity. These were *curiosity*, *fun*, and *concerns* (intrinsic), and *professional* as well as *social* (extrinsic). The goals our participants were interested in obtaining were primarily playing *games*, generating *content*, and *discovering risks* inherent to the models.

The findings also present a collection and definition of red teaming *strategies and techniques.* We categorized these into *Language strategies, Rhetoric strategies, Possible worlds strategies, Fictionalizing strategies*, and *Stratagems.* The contribution of this taxonomy is one of the first attempts at categorizing different types of red teaming attacks, and to demonstrate the techniques that they make use of. In addition to providing a sociological snapshot of the community of in-the-wild red teamers, the taxonomy can hopefully also inspire future research in and practice of creative and critical red teaming strategies.

The goal of the current work was to discover and define concepts that form LLM red teaming in the wild, at the moment when accessible, fluent-seeming, dialogue-based language models collided with broader society. Future work could include quantitative refining, verifying, and potentially testing of these concepts. The conceptual territory of how people attack LLMs is large, and the map is only just being drawn; there is much more detail to add, and there are many more pools of human knowledge to be brought to bear. For example, many successful attacks resemble manipulation and social engineering. Others use computational primitives such as matrices of floats or low-frequency tokenizer items. The intersections between LLM red teaming as it exists in the wild, and what we know about computing, security, language, and human behavior, are rich, and will likely continue to present new and intriguing results.

## Acknowledgments

We are very grateful to our participants for their time, openness, and reflections; in random order: Dan Goldstein, Vinodkumar Prabhakaran, Hyrum Anderson, Lyra Cooley, Paul Nicholas, Murat Ayfer, Djamé Seddah, Ian Ribeiro, Thomas Wood, Daniel Litt, Max Anton Brewer, Simon Willison, Sichu Lu, feddie xtzeth, Celeste Drummond, Sean Wang / swyx, Kai Greshake, Riley Goodside, Zvi Mowshowitz, Mathew Hardy, Marcin Junczys-Dowmunt, Harrison Chase, Brendan Dolan-Gavitt, Igor Brigadir, Lior Fox, Jonathan Stray, and those who preferred to remain anonymous.